\documentclass{article}

\usepackage{PRIMEarxiv}

\usepackage[utf8]{inputenc} %
\usepackage[T1]{fontenc}    %
\usepackage{hyperref}       %
\usepackage{url}            %
\usepackage{booktabs}       %
\usepackage{amsfonts}       %
\usepackage{nicefrac}       %
\usepackage{microtype}      %
\usepackage{lipsum}
\usepackage{fancyhdr}       %
\usepackage{graphicx}       %
\usepackage{subcaption}
\graphicspath{{media/}}     %
\usepackage{cleveref}

\title{Smell and Emotion: Recognising emotions in smell-related artworks}

 \author{
   Vishal Patoliya, Mathias Zinnen, Andreas Maier, Vincent Christlein \\
   Pattern Recognition Lab \\
   Friedrich-Alexander-UniversitÃ¤t \\
   Erlangen, Germany\\
}

\begin{document}
\maketitle

\makeatletter
\DeclareRobustCommand\onedot{\futurelet\@let@token\@onedot}
\newcommand{\@onedot}{\ifx\@let@token.\else.\null\fi\xspace}
\makeatother
\newcommand{\etal}{\emph{et~al\onedot}}
\newcommand{\ie}{i.\,e.,\xspace}
\newcommand{\eg}{e.\,g.,\xspace}
\newcommand{\vs}{vs\onedot}
\newcommand{\aka}{a.\,k.\,a\onedot}
\newcommand{\cf}{cf\onedot}

\section{Introduction}
Traditionally, emotions have played only a marginal role in historical research and heritage discourse.
During the twentieth century, various disciplinary turns and paradigm shifts, specifically the emergence of sensory studies~\cite{bull2006introducing} and the history of emotions~\cite{rosenwein2017history}, have broadened the perspective of humanities-based research beyond the scope of classical historiography.
Similarly, the gradual extension of the concept of cultural heritage towards intangible heritage, particularly and towards multi-sensory approaches, expands our understanding of what dimensions of culture are considered valuable to safeguard for future generations~\cite{bembibre2017smell}.
Recent approaches in computational humanities embrace this broader perspective and incorporate the recognition of abstract and subjective concepts~\cite{pandiani2023seeing}, sound~\cite{achichi2018doremus}, or smells~\cite{van2023more}.
Emotions have successfully been recognised in historical texts~\cite{rei2023detecting, massri2022harvesting} and natural images~\cite{patel2020facial, emotic_pami2019}.
Unfortunately, neural networks trained on photographic emotion recognition datasets are subject to a performance drop when applied to artworks due to the domain gap between artistic and photographic representations~\cite{hall2015cross}. 
While the WikiArt emotions~\cite{mohammad2018wikiart} and the more recent ArtEMis~\cite{achlioptas2021artemis} datasets compile emotional responses to artworks by multiple viewers, automatic recognition of emotions of the persons depicted in artworks has not yet been targeted in literature.
This work explores whether person-level emotion recognition in smell-related artworks is technically feasible. 
By focusing our study on smell-related artworks, we aim to set the stage for future research that links the extracted emotions to previously identified olfactory references~\cite{zinnen2022odor, zinnen2021see}.
With this research, we seek to contribute to the emergence of a broadened view in computational humanities, including the analysis of multiple senses, emotions, and everyday history.

\section{Method}
\subsection{Baseline Dataset}
Our approach involves training an emotion recognition network on the EMOTIon recognition in Context (EMOTIC) dataset~\cite{kosti2017emotic} and applying it to a set of smell-related artworks. 
These artworks were previously used for the detection of olfactory objects~\cite{zinnen2022odor} and gestures~\cite{zinnen2023sniffyart} in the context of the Odeuropa\footnote{\url{https://www.odeuropa.eu}} project.

The EMOTIC dataset consists of 23,571 unconstrained, natural photos with 34,320 individuals identified primarily on their perceived emotions. The dataset has images from Google, COCO~\cite{lin2015microsoft} and Ade20k~\cite{zhou2018semantic} depicting diverse contexts. 
The annotations were collected using Amazon Mechanical Turk (AMT) and agreement levels among different annotators were used in the labelling process~\cite{emotic_pami2019}. Overall, Persons are labelled with their location in the image, 26 discrete emotion categories and three continuous variables (Valence, Arousal, and Dominance). 
We labelled a subset of the Odeuropa images for quantitative evaluation using the EMOTIC labelling scheme. 
We will refer to this test set consisting of 100 images with one annotated person each as ODOR-emotions.

\subsection{Stylized Dataset Creation}
\begin{figure}[t]
    \centering
    
    \begin{subfigure}{0.3\textwidth}
        \centering
        \includegraphics[width=\linewidth]{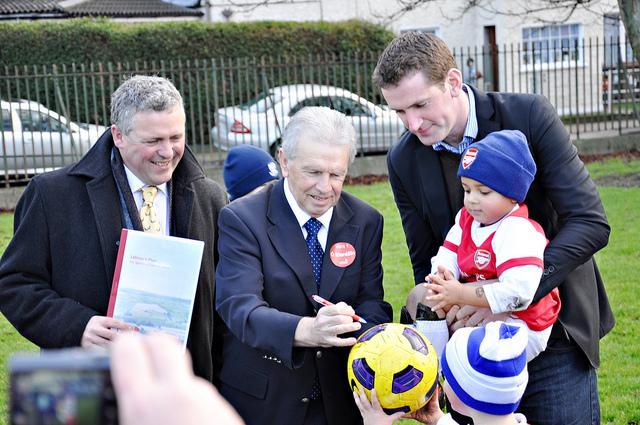}
        \caption{Content Image}
        \label{fig:image1}
    \end{subfigure}
    \hfill
    \begin{subfigure}{0.3\textwidth}
        \centering
        \includegraphics[width=\linewidth]{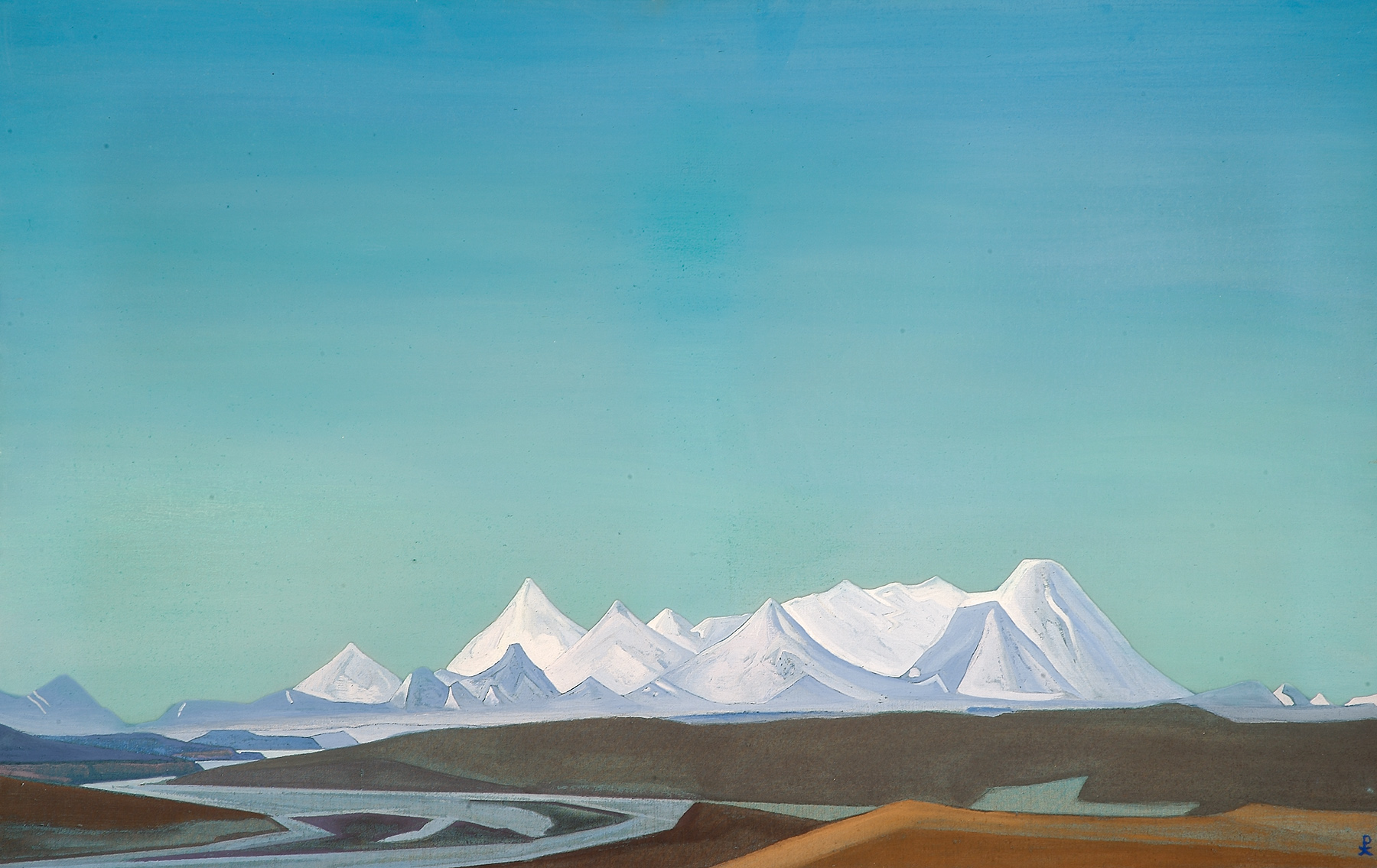}
        \caption{Style Image}
        \label{fig:image2}
    \end{subfigure}
    \hfill
    \begin{subfigure}{0.3\textwidth}
        \centering
        \includegraphics[width=\linewidth]{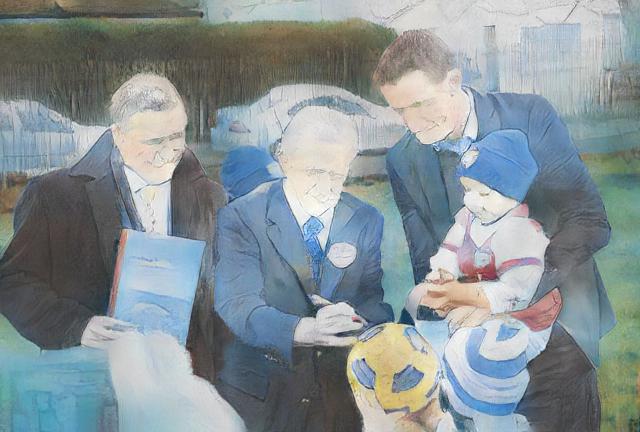}
        \caption{Style Transferred Image}
        \label{fig:image3}
    \end{subfigure}

    \caption{Example of a style transfer using internal-external learning~\cite{style}: Image (a) is the normal photographic image from the EMOTIC dataset where the style transfer will be applied. Image (b) is the artwork image from WikiArt used to get style features. Finally, Image (c) is the output image where the main body or object content is from Image (a) and the colouring effect is from Image (b).  }
    \label{fig:overall}
\end{figure}

We additionally created a stylized version of the EMOTIC dataset (EMOTIC-s) to overcome the domain gap between photographic and artistic imagery and approximate the distribution of our artistic target domain. 
To this end, we apply internal-external learning~\cite{NEURIPS2021_df535469} to the EMOTIC dataset with random artworks from the WikiArt\footnote{\url{https://www.wikiart.org/}} dataset defining the target styles (see \cref{fig:overall}).
The WikiArt dataset features images of diverse genres, such as abstract, Shan Shui, landscape, and Veduta.

\subsection{Network Architecture}
\begin{figure}[h]
\centering
\includegraphics[width=10cm]{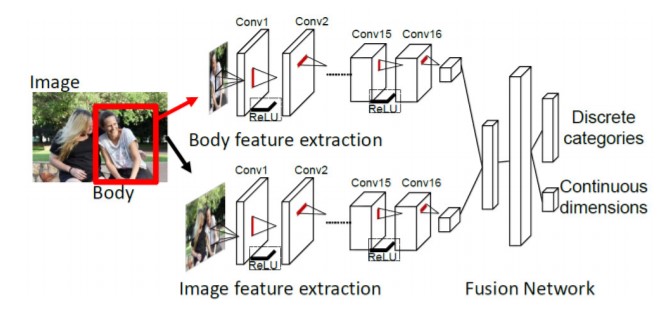}
\caption{Model architecture for Emotion Recognition in Context which uses two different branches to extract body and context features separately.
In the fusion network, both features are merged for the final prediction (Discrete and Continuous dimensions). Figure taken from \cite{emotic_pami2019} with permission from the authors.}
\label{fig:pipeline}
\end{figure}

We adopt the network architecture from Kosti et al.~\cite{emotic_pami2019}. Their convolutional neural network (CNN) model combines person-specific (body bounding box) and scene context information (entire image). The model's workflow is illustrated in \cref{fig:pipeline}, comprising three modules: body feature extraction, image (context) feature extraction, and a fusion network. 

Individual-centric attributes, such as body posture, facial expressions, and other non-verbal indicators, are the primary focus of the body feature extraction module.  It processes characteristics connected to the person's body by analyzing their visible appearance. The algorithm uses a CNN model that has been previously trained on ImageNet~\cite{ImageNet}, adjusting it to extract relevant features from each image's body bounding box. The Scene Context Feature Extraction module is tasked with understanding the broader scene-context features. The entire image is utilized as input to generate scene-related features. It employs a CNN model pre-trained on the Places dataset~\cite{zhou2017places} to extract features that capture the situational context of each image. The extracted features from both modules are combined in a fusion network to perform a detailed regression of discrete emotion categories and continuous dimensions. This network is important for interpreting emotions in artworks because it combines scene-contextual and individual-centric information to evaluate the emotional state shown in the image.

To derive features from the data, ResNet-18~\cite{He_2016_CVPR}, ResNet-50~\cite{He_2016_CVPR} and DenseNet161~\cite{Huang_2017_CVPR} neural network encoder architectures are utilized in the body and context modules and performance is evaluated for each architecture. As suggested by Kosti et al.~\cite{emotic_pami2019}, a weighted combination of two separate losses has been considered, where one loss corresponds to learning the discrete categories and the other one considers the continuous dimensions.
For the discrete categories, a weighted Euclidean loss is used, and for the continuous dimensions, a smooth $\ell^1$ loss is used.

\section{Experiments \& Results}
\subsection{Baseline Performance}
\begin{figure}[h]
    \centering
    \begin{subfigure}{0.3\textwidth}
        \centering
        \includegraphics[width=\linewidth]{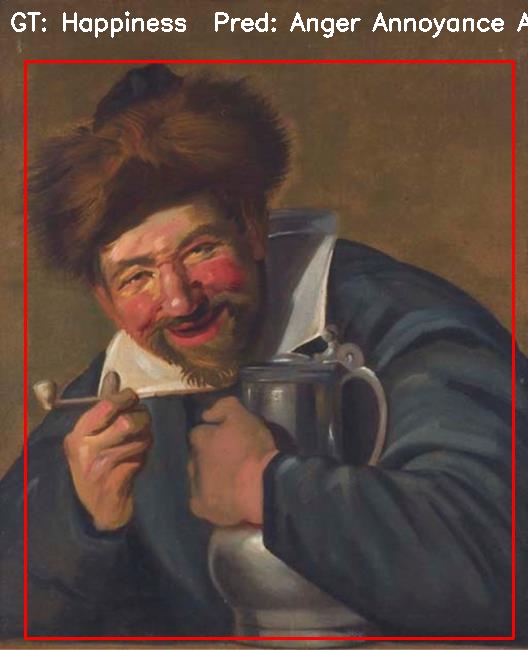}
        \caption{}
        \label{fig:image1}
    \end{subfigure}
    \qquad
    \begin{subfigure}{0.3\textwidth}
        \centering
        \includegraphics[width=\linewidth]{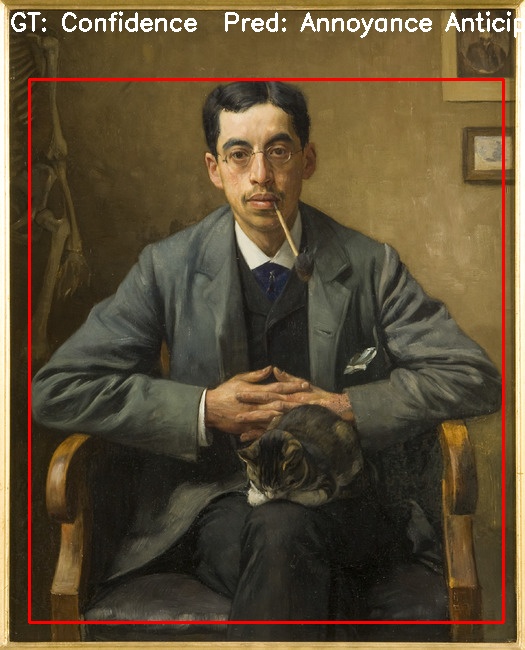}
        \caption{}
        \label{fig:image2}
    \end{subfigure}
    \hfill

    \caption{Example of misclassifications from ODOR-e dataset where Ground Truth (GT) emotion and predicted emotion (Pred) are completely different. Image credits (left to right): \textit{Merry man holding a pewter jug and a pipe}. Circle of Frans Hals. 1638--1640. \href{https://rkd.nl/en/explore/images/302304}{RKDImages(302304)}, \textit{Portrait of Arnold Aletrino}. Jan Veth. 1885. \href{https://https://rkd.nl/en/explore/images/20797}{RKDImages(20797)}.}
    \label{fig:misclassification}
\end{figure}

We retrain the architecture with ResNet-18 and ResNet-50 backbones from~\cite{emotic_pami2019}, keeping data splits, image sizes, hyperparameters and augmentations. 
We do not reproduce the configuration with a DenseNet161 due to its weaker performance reported in~\cite{emotic_pami2019}. 
~\footnote{The training is conducted using  the implementation from Abhishek Tandon:\url{https://github.com/Tandon-A/emotic/blob/master/Colab_train_emotic.ipynb}}
Kosti et al.~\cite{emotic_pami2019} evaluate the discrete prediction of the predominant emotion, and a continuous regression of the three dimensions of valence (V), arousal (A), and dominance (D).
Similarly, we evaluate our retrained models and report the average precision for the discrete emotion prediction (AP\textsubscript{EMOTIC}) and the regression error, averaged over all three dimensions (VAD\textsubscript{EMOTIC}) in \cref{tab:performance_drop}.    
The numbers are very close to what Kosti et al.~\cite{emotic_pami2019} originally reported, showcasing the reproducibility of their study. 
ResNet-50 stands out as the superior model when EMOTIC and ODOR-e datasets are taken into consideration.
However, if we apply the models on our test set consisting of historical artworks, we see a drop in classification performance (measured by AP\textsubscript{ODOR-e}) and an increase in regression error (measured by VAD\textsubscript{ODOR-e}). Examples of misclassifications in the ODOR-e dataset were shown in \cref{fig:misclassification}.

\begin{table}[h]
 \caption{Comparison of average precision for predominant emotion prediction (AP) and mean error in valence arousal dominance dimensions (VAD) of models trained on the EMOTIC. 
    Both models exhibit a decrease in classification performance and an increase in regression error for the continuous dimensions when evaluating on artworks compared to the evaluation on the EMOTIC test set.}
    \centering
    \begin{tabular}{lcccc}
        \toprule
        Model & AP\textsubscript{EMOTIC}$\uparrow$ &  AP\textsubscript{ODOR-e}$\uparrow$ & VAD\textsubscript{EMOTIC}$\downarrow$ & VAD\textsubscript{ODOR-e}$\downarrow$ \\
        \midrule
         ResNet-18 & 25.61 &  13.25 & 0.97 & 1.89 \\
         ResNet-50 & 26.13 & 13.33 & 0.96 &  1.91  \\
         \bottomrule \\
    \end{tabular}   
    \label{tab:performance_drop}
\end{table}

\subsection{Improvements on Baseline Performance}
To overcome this performance degradation, we adapt the original configurations in two ways:
\begin{enumerate}
    \item We replace the backbone of the context branch, which was originally pretrained with Places365~\cite{zhou2017places} with an ImageNet-pretrained backbone.
    \item We increase the size of the person crops that are being fed to the body feature extraction branch of the network to incorporate more detailed local features.
\end{enumerate}

Results of these alterations are reported in \cref{tab:hyperparams}. 
Changing the pre-training scheme improves performance when evaluated on EMOTIC.
Evaluated on ODOR-e, we observe a decrease in classification accuracy while at the same time reducing the VAD regression error.
Increasing the size of the person crops, however, increases the classification performance on both datasets while the regression error stays roughly the same.
When both techniques are combined, the classification performance on EMOTIC is slightly improved while being decreased for ODOR-e.
Since we are mostly interested in predominant emotion classification on ODOR-e, we conducted further experiments with the increased image size of 224. 

\begin{table}[h]
 \caption{Ablation study of Hyperparameter tuning, models trained with ImageNet weights in context branch (INW) or increased body image size to 224 (224B) or in the combination of both. Models with increased body image size perform better than models trained only with INW in the context branch. }
    \centering
    \begin{tabular}{lcccccc}
    \toprule
        Model & INW & 224B &AP\textsubscript{EMOTIC}$\uparrow$ &  AP\textsubscript{ODOR-e}$\uparrow$ & VAD\textsubscript{EMOTIC}$\downarrow$ & VAD\textsubscript{ODOR-e}$\downarrow$ \\
         \midrule
         ResNet-50 & & & 26.13 & 13.33 & 0.96 &  1.91  \\
         ResNet-50 &\checkmark &  & 26.50 & 10.22 & 0.96 & 1.81\\
         ResNet-50 & & \checkmark & 26.90 & 13.49 & 0.95 & 1.92\\
         ResNet-50 &\checkmark & \checkmark & 27.04 & 13.33 & 0.95 & 1.88\\
         \bottomrule \\
    \end{tabular}   
    \label{tab:hyperparams}
\end{table}

\subsection{Impact of Style Transfer}
To further increase model performance on the artistic person representations in ODOR-e, we experiment with a stylized version of the EMOTIC dataset (EMOTIC-s). 
\Cref{tab:my_label} compares the performance of ResNet-18 and ResNet-50 models when trained on EMOTIC and EMOTIC-s datasets, respectively.

\begin{table}[h]
\caption{Ablation study comparing model performance when trained on EMOTIC vs when trained on EMOTIC-s dataset. Both configurations are evaluated on EMOTIC-s and ODOR-e datasets.}
    \centering
    \begin{tabular}{llcccc}
    \toprule
    Model & Train Set & AP\textsubscript{EMOTIC-s}$\uparrow$ & AP\textsubscript{ODOR-e} $\uparrow$
        & VAD\textsubscript{EMOTIC-s}$\downarrow$  & VAD\textsubscript{ODOR-e}$\downarrow$ \\
    \midrule
        ResNet-18 + 224B & EMOTIC & - & 15.05 & - & 1.97\\
        ResNet-18 + 224B & EMOTIC-s & 23.45 & 12.28 & 0.99 & 1.94\\
        ResNet-50 + 224B & EMOTIC & - & 13.49 & - & 1.92\\
        ResNet-50 + 224B & EMOTIC-s & 23.59 & 13.93 & 0.98 & 1.86\\
        \bottomrule \\
    \end{tabular}
    
    \label{tab:my_label}
\end{table}

While we observe a slight improvement in the performance of ResNet-50 models trained on the EMOTIC-s dataset, the performance for the strongest ResNet-18 model deteriorates drastically. 
However, the performance increase for the larger backbone motivates us to further explore the potential of style transfer in future work.
Unfortunately, neither the models with a ResNet-18 backbone nor those with a ResNet-50 backbone exhibit improved performance compared to their counterparts trained on the unmodified EMOTIC dataset. 
We hypothesize that the style transfer leads to a loss of crucial information about the facial features of depicted persons. 
This is exemplified in the transition from \cref{fig:image1} to \cref{fig:image3}. 
Developing conditioned style-transfer methods that ensure consistency in the facial expression of depicted persons could mitigate this issue and provide a promising angle for future work.

\section{Conclusion}
The study shows the feasibility of recognizing emotions in individuals portrayed in artworks and highlights the importance of contextual information for identifying these emotions.
However, we observe a performance decrease compared to emotion recognition in natural images. 
We employed customized Convolutional Neural Network (CNN) models, pre-trained on the photographic EMOTIC dataset, and achieved modest improvements through hyperparameter tuning. Furthermore, we explored the potential of style transfer to overcome the domain gap between natural and artistic representations.
The recognition performance in artworks has substantial room for improvement, given the complexity and subjectivity of the problem.
However, the proposed method presents a first exploration of the domain, enabling further research in emotion recognition in artworks.
Future studies might focus on extending the dataset to enable fine-tuning in the artistic target domain. 
Similarly, improving style transfer methods might retain features crucial for emotion recognition.
Ultimately, we hope this work encourages interdisciplinary scholars to investigate the interplay between olfactory and emotional dimensions of artworks.
\section*{Acknowledgments}
This paper has received funding from the Odeuropa EU H2020 project under grant agreement No.\ 101004469. 
We gratefully acknowledge the donation of the NVIDIA corporation of two Quadro RTX 8000 that we used for the experiments.

\bibliographystyle{unsrt}  
\bibliography{references}

\end{document}